\documentclass[letterpaper, 10 pt, journal]{ieeeconf} 
\usepackage[left=0.75in,right=0.75in,top=0.75in,bottom=0.75in]{geometry}

\usepackage{graphics} % for pdf, bitmapped graphics files
\usepackage{graphicx}   % To include figures
\usepackage{amsmath} % assumes amsmath package installed
\usepackage{amssymb}  % assumes amsmath package installed
\usepackage{mathtools, cuted}
\usepackage{interval}
\usepackage{nicefrac}
\usepackage[noadjust]{cite}
\usepackage{breqn}
\usepackage{color, colortbl}
\usepackage{derivative}
\usepackage{siunitx}
\usepackage{multirow}
\usepackage{hyperref}
\usepackage{cleveref}

\renewcommand{\vec}[1]{\mathbf{\boldsymbol{#1}}}

\usepackage{marginnote}

\usepackage[dvipsnames]{xcolor}

\providecommand{\Review}[3]{#2}

\title{Refining Motion for Peak Performance: \Review{Optimizing}{Identifying Optimal}{6-6} Gait \\ Parameters for Energy-Efficient Quadrupedal \Review{Robots Locomotion}{Bounding}{6-6}}
 
\author{Yasser G. Alqaham$^*$, Jing Cheng$^*$, and Zhenyu Gan
\thanks{All authors are with the Department of Mechanical and Aerospace Engineering, Syracuse University, Syracuse, NY 13244 \texttt{\{ygalqaha, jcheng13, zgan02\}@syr.edu}.}
\thanks{This work was supported by a startup fund from the Syracuse University.}
\thanks{$^*$The authors contribute equally to this paper.}
}

\begin{document}

\maketitle

\thispagestyle{empty}
\pagestyle{empty}

\begin{abstract}
Energy efficiency is a critical factor in the performance and autonomy of quadrupedal robots. While previous research has focused on mechanical design and actuation improvements, the impact of gait parameters on energetics has been less explored. In this paper, we hypothesize that gait parameters, specifically duty factor, phase shift, and stride duration, are key determinants of energy consumption in quadrupedal locomotion. To test this hypothesis, we modeled the Unitree A1 quadrupedal robot and developed a locomotion controller capable of independently adjusting these gait parameters. Simulations of bounding gaits were conducted in Gazebo across a range of gait parameters at three different speeds: low, medium, and high. Experimental tests were also performed to validate the simulation results. The findings demonstrate that optimizing gait parameters can lead to significant reductions in energy consumption, enhancing the overall efficiency of quadrupedal locomotion. This work contributes to the advancement of energy-efficient control strategies for legged robots, offering insights directly applicable to commercially available platforms.
\end{abstract}

\section{Introduction}
Quadrupedal robots have garnered significant attention in recent years due to their potential to navigate complex terrains and perform tasks that are challenging for wheeled or tracked robots \cite{Raibert2008, Hyun2014}. \Review{However, there is a notable lack of research focusing on the energetics of locomotion in these systems, which is crucial for enhancing their practicality, operational endurance, and overall efficiency \cite{Belter2016, Semini2011}. Energy efficiency directly impacts the work time and applicability of legged robots in real-world scenarios.}{Energy efficiency is crucial for the practicality, operational endurance, and overall effectiveness of legged robots in real-world scenarios. While there is existing research on locomotion energetics in these systems, it remains limited and insufficient to fully optimize their performance \cite{Semini2011, Xi2015, Belter2016, Kolvenbach2018}}{1-1}. Studies in animal locomotion have demonstrated that gait parameters have a substantial influence on energy economy \cite{Alexander1989, Hoyt1981}. For instance, animals naturally adjust gait parameters such as duty factor, phase shift, and stride duration to optimize energy consumption during movement \cite{Grillner1981, Heglund1985}. 
% Understanding and applying these principles to robotic systems could lead to significant improvements in their energy efficiency and functionality.
% \emph{Gait parameters} such as phase shift, duty factor, and stride duration fundamentally affect the dynamics and energetics of quadrupedal locomotion \cite{Hilton2018, Abernethy2016}. 
\emph{Duty factor} is the proportion of the gait cycle during which a foot is in contact with the ground (normalized by the total stride time) \cite{Grillner1981}. Animals adjust their duty factor to maintain stability and optimize energy use; for instance, at higher speeds, quadrupeds often decrease their duty factor, resulting in shorter ground contact time and longer aerial phases \cite{Hildebrand1965}. Cheetahs, known for their high-speed pursuits, exhibit low duty factors during sprints to maximize speed while balancing energy expenditure \cite{Hudson2012}. 
\emph{Phase shift} refers to the temporal offset between the movements of different legs during a gait cycle \cite{Alexander1984}. In quadrupeds, altering the phase shift can transition the gait from a walk to a trot or gallop. For example, horses change their gait from trot to gallop by adjusting the phase shift between their legs, which affects their speed and energy efficiency \cite{Hoyt1981, Heglund1985}. 
\emph{Stride duration} is the time taken to complete one full gait cycle \cite{Abernethy2016}. Animals alter their stride duration based on desired speed and energetic efficiency. Elephants, for example, use longer stride durations at slower speeds to minimize energy expenditure \cite{Hutchinson2006}. In contrast, smaller animals like mice have shorter stride durations due to their anatomical constraints \cite{Heglund1974}.

Most robotic systems, however, employ fixed gait parameters designed for all tasks or speed ranges rather than dynamically adjusting them as animals do. This approach can lead to suboptimal energy usage, particularly when robots need to operate across varied terrains or speeds. Many legged robots rely on predefined gaits that fail to adjust stride duration or duty factor in response to changes in speed or load, resulting in increased energy expenditure during operation \cite{Raibert1986}.
Previous studies have investigated gait optimization for speed and stability \cite{Kim2011, Karssen2011}, but there is a scarcity of research focusing specifically on energy consumption related to gait parameters. For instance, Geyer et al. \cite{Geyer2006} analyzed compliant leg behaviors but did not extensively explore the impact of varying phase shifts or duty factors on energy efficiency. Similarly, Fei et al. \cite{Fei2018} concentrated on mechanical aspects rather than control strategies for energy optimization. \Review{}{Silva et al. \cite{Silva2006} examined how a quadrupedal robot's performance is influenced by general locomotion parameters, such as step length, body height, and forward speed. However, these parameters are not specific to gait mechanics but rather represent broader aspects of locomotion}{1-1}. Robots like Boston Dynamics' Spot or MIT's Cheetah use different gaits, but their efficiency is often hindered by the rigidity of their gait parameter settings \cite{Park2017, Katz2019}. In robotics, a fixed phase shift and duty factor may limit the robot's ability to adapt to different speeds or terrains, potentially leading to increased energy consumption. And adjusting stride duration can help match the robot's natural frequency to that of its mechanical structure, reducing unnecessary energy consumption due to mechanical impedance mismatches.

Energy efficiency remains a limiting factor for the deployment of quadrupedal robots in long-duration tasks and remote operations \cite{Bledt2018}. Despite advances in hardware, there is a need for control strategies that optimize energy usage without compromising performance. We hypothesize that gait parameters, specifically duty factor, phase shift, and stride duration, are key determinants of energy consumption in locomotion. By modifying these parameters, energy consumption during movement can be directly influenced, resulting in more efficient locomotion. We propose that these parameters should not remain fixed but should dynamically adjust according to changes in speed, load, or terrain. Just as animals adapt their gait to minimize energy expenditure across various conditions, legged robots should utilize real-time adaptive strategies to optimize their cost of transport.
The primary objective of this study is to examine how variations in gait parameters affect the energetics of quadrupedal locomotion. To achieve this, we developed a dynamic model of the Unitree A1 robot that accurately represents its physical properties and \Review{designed}{utilized}{1-3} a locomotion controller capable of independently adjusting duty factor, phase shift, and stride duration. Simulations were conducted in Gazebo to analyze energy consumption across different gait parameters at low, medium, and high speeds, and experimental validations were performed to \Review{confirm}{verify}{1-3} the simulation results.

Our \emph{contributions} are as follows: we demonstrate that optimizing gait parameters can lead to \Review{significant}{noticeable}{1-2} reductions in energy consumption \Review{}{in quadrupedal bounding gaits}{1-2}; we provide a systematic approach to gait parameter manipulation for energy efficiency; and we \Review{validate}{compare}{1-3} simulation findings with experimental results on a real robot platform.
The rest of the paper is organized as follows: Section~\ref{sec:methods} provides an in-depth explanation of the robot's dynamic model, the structure of the proposed control framework, and the cost of transport calculations. In Section~\ref{sec:results}, we evaluate the impact of three gait parameters (duty factor, phase shift, and stride time) on the robot's energy consumption and performance, focusing on a bounding gait with two flight phases. Lastly, Section~\ref{sec:conclusions} summarizes the key findings and conclusions of the study.

%%%%%%%%%%%%%%%%%%%%%%%%%%%%%%%%%%%%%%%%%%%%%%%%%%
%%%%%%%%%%%%%%%%%%%%%%%%%%%%%%%%%%%%%%%%%%%%%%%%%%
\section{Methods}
\label{sec:methods}
This section provides a detailed description of the A1 quadrupedal robot's dynamics model used in this study, outlines the proposed framework for the locomotion control with adaptable parameters, and includes the metric that is used to evaluate the energy economy.

\subsection{Robot Model}
\label{sec:Robot Model}
\begin{figure}[tbp]
\centering
\includegraphics[width=1\columnwidth]{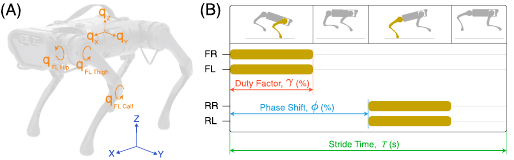}
\caption[Gait Generation]{This figure illustrates the gait parameters in one stride for the bounding gait with two flight phases. The yellow stripes represent the stance time of the front and rear legs in their respective phases, which can be expressed by the duty factor (shown by the red arrow) as a percentage of the stride duration. The blue arrow indicates the phase shift, which is the time difference between the rear legs touchdown and front legs touchdown. The stride duration is depicted by the green arrow. FR, FL, RR, RL, correspond to front-right leg, front-left leg, rear-right leg, and rear-left leg, respectively.} 
\label{fig:GaitGeneration}
\vspace{-2mm}
\end{figure}

%Figure : Kinematics Diagram
% \begin{figure}[tbp]
% \centering
% \includegraphics[width=1\columnwidth]{Fig_2.pdf}
% \caption[Kinematic Model]{Figure (A) shows the kinematic model of A1 robot along with configuration variables. Each leg has three joints namely hip, thigh, and calf. Figure (B) demonstrates the footfall patterns for pronking and three bounding gaits. \Zhenyu{This figure needs some work, we need to show how different models are been used for control the front/rear hip motions.}} 
% \label{fig:Kinematics Diagram}
% \vspace{-2mm}
% \end{figure}
The \Review{}{floating-base model of the}{2-1} Unitree A1 quadrupedal robot, shown in Fig.~\ref{fig:GaitGeneration}(A), features 18 degrees of freedom (DOF), with each leg comprising three revolute joints powered by electric motors, and the torso \Review{(floating base)}{}{2-1}possessing six DOF representing its position and orientation in space. The robot's torso position is described using Cartesian coordinates $\vec{q}_{\text{pos}} = [q_x, q_y, q_z]^\intercal$ relative to the inertial frame, while its orientation is specified by Euler angles $\vec{q}_{\text{ori}} = [q_{\text{roll}}, q_{\text{pitch}}, q_{\text{yaw}}]^\intercal$. Each leg's joint angles are represented as $\vec{q}_{i,\text{leg}} = [q_{i,\text{hip}}, q_{i,\text{thigh}}, q_{i,\text{calf}}]^\intercal$, indicating the relative angular positions of the hip to the torso, the thigh to the hip, and the calf to the thigh, following the right-hand rule. The legs are labeled as $i \in \{\text{FR}, \text{FL}, \text{RR}, \text{RL}\}$, corresponding to front-right, front-left, rear-right, and rear-left, respectively. The full configuration vector of the robot is $\vec{q} = [\vec{q}_{\text{pos}}^\intercal, \vec{q}_{\text{ori}}^\intercal, \vec{q}_{\text{FR,leg}}^\intercal, \vec{q}_{\text{FL,leg}}^\intercal, \vec{q}_{\text{RR,leg}}^\intercal, \vec{q}_{\text{RL,leg}}^\intercal]^\intercal \in \mathbb{R}^{18}$.
The dynamics of the robot's floating-base model are derived using the Euler-Lagrange equations, resulting in the equation of motion:
\begin{equation}
\label{eq:EOM}
   \vec{M}(\vec{q})\ddot{\vec{q}} + \vec{C}(\vec{q},\dot{\vec{q}}) + \vec{G}(\vec{q}) = \vec{S}\vec{\tau}  + \sum_{i}\vec{J}_{i}^\intercal(\vec{q})\vec{\lambda}_i,
\end{equation}
where $\vec{M}(\vec{q}) \in \mathbb{R}^{18 \times 18}$ is the mass-inertia matrix, $\vec{C}(\vec{q},\dot{\vec{q}}) \in \mathbb{R}^{18}$ represents the Coriolis and centrifugal force vector, and $\vec{G}(\vec{q}) \in \mathbb{R}^{18}$ is the gravitational force vector. The vector $\vec{\tau} \in \mathbb{R}^{12}$ contains the joint torques, and the selection matrix {\footnotesize $\vec{S} = \begin{bmatrix}\vec{0}_{6 \times 12} \\ \vec{I}_{12 \times 12}\end{bmatrix}$} associates motor torques with their respective joints. The term $\vec{J}_{i}(\vec{q})$ is the contact Jacobian matrix for the $i$-th foot \Review{}{(see the detailed definition in the Appendix)}{8-1}, and $\vec{\lambda}_{i}$ denotes the ground reaction force (GRF) at that foot. The summation $\sum_{i}\vec{J}_{i}^\intercal(\vec{q})\vec{\lambda}_i$ projects the GRFs from the contact points into the joint space.

\begin{figure*}[tbp]
\centering
\includegraphics[width=2\columnwidth]{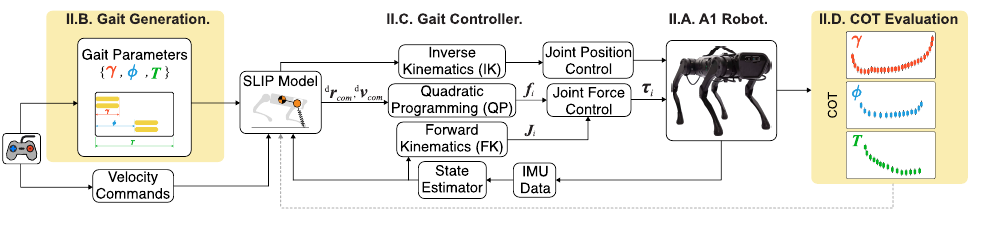}
\caption[ControlDiagram]{This control diagram illustrates the quadrupedal robot's gait generation and control system. The Gait Generation block determines the key gait parameters: duty factor ($\gamma$), phase shift ($\phi$), and stride time ($T$). The Gait Controller utilizes the SLIP Model, along with a quadratic programming (QP) solver, inverse kinematics (IK), and forward kinematics (FK), to compute joint positions and ground reaction forces (GRF), based on velocity commands and state estimation data. COT Evaluation block assesses the cost of transport, optimizing energy efficiency for different values of $\gamma$, $\phi$, and $T$} 
\label{fig:ControlDiagram}
\vspace{-2mm}
\end{figure*}

\subsection{Gait Generation}
For gait generation, we employ a periodic method that uses two parameters for each leg: the duty factor and the phase shift. The duty factor $\gamma_i$ represents the stance time of each leg as a percentage of the total stride duration $T$, while the phase shift $\phi_i$ determines the timing offset of each leg's movement relative to a reference leg, also expressed as a percentage of $T$. Adjusting these parameters collectively shapes the footfall patterns that define different gaits. Since most gaits are inherently periodic, tuning these timing parameters within each cycle allows for clear differentiation between gaits. In general, specifying the gait of a quadrupedal robot requires defining eight parameters: the duty factor and phase shift for each of the four legs. Another important parameter is the stride duration $T$, which, while not affecting the footfall patterns, influences the gait cycle frequency and consequently the robot's speed. Expressing the duty factors and phase shifts as proportions of $T$ ensures that the footfall patterns remain consistent even when the gait frequency changes.
In this study, we focus on a specific \emph{bounding gait} characterized by dual flight phases, where the front and rear legs move in pairs and alternate between stance and swing phases \cite{alqaham2024energy}. \Review{}{Bounding, unlike trotting, inherently allows for greater aerial phases, which can lead to reduced ground contact losses and improved energy efficiency at higher speeds. Moreover, bounding has been observed in agile quadrupedal locomotion in both animals and robots, making it a compelling choice for studying energetically optimal gait patterns.}{1-5} Since the legs in each pair move synchronously \Review{}{in bounding}{}, and we assume the front and rear legs share the same duty factor, the eight parameters required to define a general gait are reduced to just three: the duty factor $\gamma$ for the front and rear legs, the phase shift $\phi$ between them, and the total stride time $T$, as illustrated in Fig.~\ref{fig:GaitGeneration}(B).

\subsection{Bounding Gait Control}
\Review{We adopted the control approach outlined in \cite{cheng2024}, integrating two key techniques. First, we model the front or rear legs as a Spring-Loaded Inverted Pendulum (SLIP) during their stance phases to generate the desired accelerations of the robot's center of mass (COM).}{This study adopts a novel control approach developed by the authors \cite{cheng2024}, integrating two key techniques. First, instead of modeling the torso as a Spring-Loaded Inverted Pendulum (SLIP) system \cite{gan2017SLIP}, the front or rear legs are individually treated as SLIP systems during their stance phases. This enables direct generation of the desired accelerations for the robot's center of mass (COM).}{1-3} This \Review{model}{technique}{1-3} emulates natural leg elasticity and captures the inherent oscillations of the torso. Second, we apply a \Review{}{standard}{1-3} \Review{single rigid body}{Single Rigid Body (SRB)}{} approximation to simplify the system dynamics, allowing for efficient computation of the GRFs needed to achieve the desired motion. This method facilitates real-time control by producing desired COM accelerations at a high refresh rate, with the overall control structure depicted in Fig.~\ref{fig:ControlDiagram}.
In bounding gaits, the robot supports itself entirely on either the front or rear legs during their respective stance phases. This reliance on a single pair of legs per phase inevitably causes substantial rotations of the torso in the sagittal plane. Instead of forcing the torso to maintain a fixed pitching angle during these phases, our controller focuses on regulating movement through the hip joints of the stance legs, allowing the torso to rotate freely. We use the SLIP model to approximate the vertical position of the hip joints during the front or rear stance phases, leading to the following approximated sinusoidal dynamics:
\begin{equation}
\label{eq:hip_motion}
\begin{aligned}
r_{hip,z} = & -\left(a_1 + a_2 \, v_{com,x} \right) \sin\left(\frac{\pi}{T \gamma} t_{s}\right) \\ 
&+ \left( a_3 - a_4 v_{com,x} \right),
\end{aligned}
\end{equation}
where $v_{com,x}$ is the horizontal speed of the COM, $T$ is the total stride duration, $\gamma$ is the duty factor, $t_s$ is the time elapsed since the beginning of the current stance phase, and $a_1$, $a_2$, $a_3$, and $a_4$ are parameters determined through curve fitting to accommodate the velocity dependencies indicated by the SLIP model. This equation accounts for the increasing angle of attack and spring compression associated with higher forward speeds. The desired motion of the COM can then be derived using the kinematic equations ${}^{\text{d}}\vec{r}_{com} = {}^{\text{d}}\vec{r}_{hip} + \vec{r}_{hip/com}$ and ${}^{\text{d}}\vec{v}_{com} = {}^{\text{d}}\vec{v}_{hip} + \vec{\omega} \times \vec{r}_{hip/com}$, where ${}^{\text{d}}(\cdot)$ denotes desired reference values, $\vec{r}_{hip/com}$ is the relative position between the hip and the COM, and $\vec{\omega}$ is the angular velocity of the torso.

After determining the desired COM position and velocity, the Quadratic Program (QP) computes the GRFs needed to guide the robot along the specified trajectory. Modeling the robot as a floating single rigid body with GRFs applied at the foot contact points, the QP optimization identifies the optimal GRFs by minimizing a cost function that balances the magnitude of the GRFs and their deviation from the previous iteration while satisfying \Review{dynamic and friction}{SRB dynamics and foot contact friction}{8-2} constraints. The cost function is defined as:
\begin{align}
\label{eq:QP_cost_function}
J &= \frac{1}{2} \boldsymbol{\lambda}^{\text{T}}\left(\boldsymbol{A}^{\text{T}} \boldsymbol{W}_1 \boldsymbol{A} + \alpha \boldsymbol{W}_2 + \beta \boldsymbol{W}_3\right) \boldsymbol{\lambda} \\ \notag
&+ \left(-\boldsymbol{b}_d^{\text{T}} \boldsymbol{W}_1 \boldsymbol{A} - \boldsymbol{\lambda}_{\text{prev}}^{\text{T}} \beta \boldsymbol{W}_3\right) \boldsymbol{\lambda},
\end{align}
where $\boldsymbol{\lambda} \in \mathbb{R}^{12}$ aggregates the GRFs at all feet, $\boldsymbol{A}$ \footnote{ $\boldsymbol{A} =  \begin{bmatrix} \boldsymbol{1} & \boldsymbol{1} & \boldsymbol{1} & \boldsymbol{1} \\ {\left[\vec{r}^{\text{FR}}\right]_{\times}} & {\left[\boldsymbol{\vec{r}}^{\text{FL}}\right]_{\times}} & {\left[\boldsymbol{\vec{r}}^{\text{RR}}\right]_{\times}} & {\left[\boldsymbol{\vec{r}}^{\text{RL}}\right]_{\times}} \end{bmatrix}$, $\vec{r}^i$ is the vector from the COM to the $i$-th foot contact point, and $[\cdot]_{\times}$ denotes the skew-symmetric matrix of a vector.}
maps GRFs to the wrench on the COM, $\boldsymbol{b}_d$ \footnote{$\boldsymbol{b}_d = \begin{bmatrix} m(\dot{\boldsymbol{v}}_{com} - \boldsymbol{g}) \\ \vec{R} \boldsymbol{I} \vec{R}^{\text{T}} \dot{\boldsymbol{\omega}} \end{bmatrix}$. $\vec{R}$ is the rotation matrix that transforms coordinates from the body-fixed coordinate frame to the inertial frame. $m$ denotes the mass of the simplified model. $\boldsymbol{I}$ refers to the inertia tensor of the simplified torso.} 
represents the desired wrench on the COM, $\boldsymbol{W}_1$, $\boldsymbol{W}_2$, and $\boldsymbol{W}_3$ are weighting matrices, $\alpha$ and $\beta$ are scalar weighting factors, $\boldsymbol{\lambda}_{\text{prev}}$ is the GRFs vector from the previous iteration. To ensure the GRFs are within the friction cone, we impose linear constraints: $\vec{M}_{\mu} \boldsymbol{\lambda} \geqslant \mathbf{0}$, where $\vec{M}_{\mu}$ incorporates the static friction coefficient, chosen to be 0.6. This approach has been effectively utilized by Unitree Robotics in the A1 robot to execute various gaits \cite{Unitree}.

Once the desired GRFs are computed, we convert them into joint torques for each leg. During the stance phase, the desired hip positions and velocities, ${}^{\text{d}}\vec{r}_{hip}$ and ${}^{\text{d}}\vec{v}_{hip}$, are obtained and converted into joint positions and velocities using inverse kinematics (IK). The joint torques for the $i$-th leg are then calculated as $\vec{\tau}_{i} = \vec{J}_{i}^\intercal(\vec{q})\vec{\lambda}_i$, where $\vec{J}_{i}(\vec{q})$ is the Jacobian of the $i$-th leg. During the swing phase, the desired joint angles and velocities are derived from foot trajectories, following foot placement strategies inspired by Raibert's approach \cite{Raibert1986}.

\subsection{Cost of Transport (COT)}
Minimizing energy consumption is crucial for maximizing the operational time of robots, especially those relying on battery power. The efficiency of a robot's locomotion is commonly evaluated using the COT, which quantifies the energy required to move a unit mass over a unit distance. COT is influenced by factors such as the robot's morphology, mass, terrain properties, speed, gait type, and control strategy. \Review{We define the work-based COT as the dimensionless quantity:}{We utilized the work-based COT defined in \cite{alqaham2024energy} as the dimensionless quantity:}{8-3}
\begin{equation}
\label{eq:COT}
\text{COT} = \frac{\int_{t_o}^{t_f} \sum_{r=1}^{12} | \tau_r \cdot \dot{q}_{r}| \, dt}{mg \left( q_{x}(t_f) - q_{x}(t_o) \right)},
\end{equation}
where $t_o$ and $t_f$ are the start and end times of the stride, $\tau_r$ and $\dot{q}_{r}$ are the torque and angular velocity of the $r$-th joint, $m$ is the total mass of the robot, $g$ is the acceleration due to gravity, and $q_{x}(t_f) - q_{x}(t_o)$ is the distance traveled in the forward ($x$) direction. The absolute value accounts for both positive and negative mechanical work.

%%%%%%%%%%%%%%%%%%%%%%%%%%%%%%%%%%%%%%%%%%%%%%%%%%
\section{Results}
\label{sec:results}
This section presents the comprehensive findings from our simulation and experimental studies on how gait parameters specifically duty factor, phase shift, and stride duration affect the energy consumption of the Unitree A1 quadrupedal robot. The simulation results, conducted using ROS \cite{ROS} and Gazebo \cite{koenig2004design}, illustrate the impact of varying each gait parameter across different speeds on the COT. 
During both simulation and experimental trials, the Unitree A1 quadrupedal robot was initially positioned in a stance state and gradually accelerated to the target speed. Once the desired speed was achieved, gait parameters were systematically adjusted until the robot reached a steady-state gait. Throughout this process, data including joint velocities and torques were continuously recorded for approximately 30 gait cycles to ensure comprehensive data collection. In instances where the robot failed to attain a steady state, resulting in unstable locomotion, the corresponding data sets for those specific gait parameter configurations were omitted from the analysis. The code and video animations of the simulation results can be found in \footnote{\href{https://github.com/DLARlab/OptimizingGaitParams.git}{https://github.com/DLARlab/OptimizingGaitParams.git}}.
\begin{figure*}[h]
\centering
\includegraphics[width=2\columnwidth]{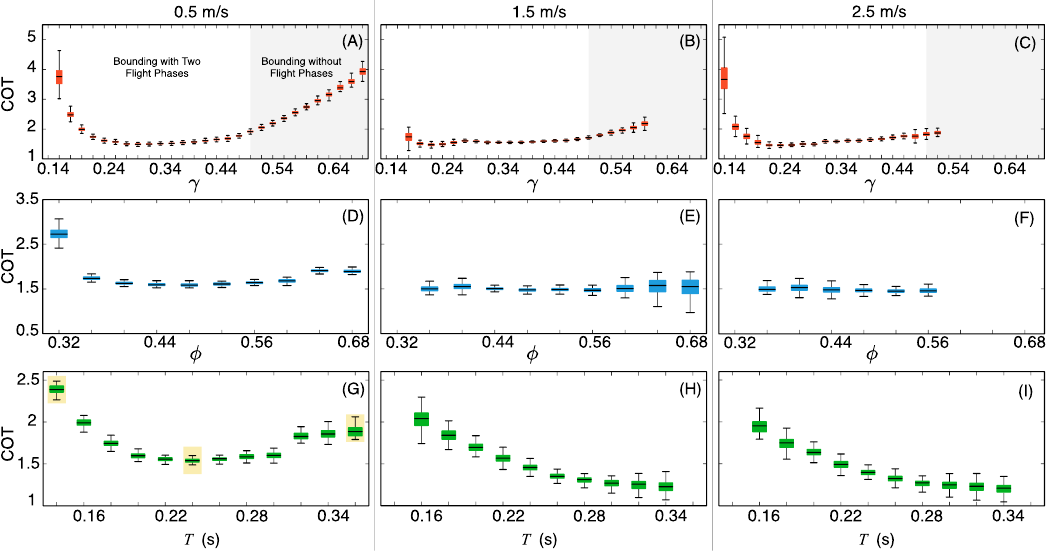}
\caption[COT]{This figure shows the cost of transport (COT) as a function of three key gait parameters: duty factor (red), phase shift (blue), and stride duration (green), for the bounding gait at three speeds (0.5 m/s, 1.5 m/s, and 2.5 m/s). \Review{}{Each box shows the median as the central mark, with the bottom and top edges representing the 25th and 75th percentiles, respectively, and the whiskers span the range of data points excluding outliers.}{8-5} The analysis indicates that higher duty factors, particularly in the context of bounding with two flight phases, and longer stride durations significantly lower the COT at all speeds, while variations in phase shift have more subtle effects on energy efficiency. The three highlighted cases in (G) are discussed in detail in Section~\ref{sec:results}.B and Fig. 4.} 
\label{fig:COT}
\vspace{-2mm}
\end{figure*}

\subsection{Optimal COT in Simulation}
To understand how the gait parameters influence the energy consumption and the resulting COT, we conducted a sensitivity test for the three key gait parameters at three distinct speeds 0.5 m/s, 1.5 m/s, and 2.5 m/s respectively. As shown in Fig.~\ref{fig:COT}, each column in the figure corresponds to a specific speed, while each row illustrates the relationship between one of these parameters and the COT. % The trends observed across the three speeds provide insights into how changes in phase shift $\phi$, duty factor $\gamma$, and stride duration $T$ affect energy consumption, enabling optimization of the robot’s gait to minimize energy expenditure.
In these simulations, the default duty factor ($\gamma$), phase shift ($\phi$), and stride duration ($T$) are selected as 0.22, 0.50, and 0.22 s, respectively, based on the trajectory optimization results from our prior work in \cite{alqaham2024energy}.

\subsubsection{0.5 m/s} At low speeds, as shown in Fig.~\ref{fig:COT}(A), the COT decreases sharply from approximately 4 at a duty factor $\gamma$ of 0.16 to 1.8 at 0.22, after which it stabilizes. When the duty factor is extremely small, the stance phase becomes shorter, requiring larger GRFs to maintain balance, which increases the challenge of stabilizing the robot and leads to higher energy consumption. As the duty factor increases beyond 0.50, the robot transitions into a bounding gait without a flight phase, causing a significant increase in COT. With a fixed stride time, larger duty factors result in shorter flight phases, which in turn increases energy consumption due to the greater effort required to accelerate and decelerate the legs. A large duty factor beyond 0.68 does not provide enough time in the swing phase of each leg to reach the desired position, ultimately leading to failure. This analysis indicates that duty factors in the range of 0.22 to 0.30 are optimal for maintaining a consistently low energy cost.

As depicted in Fig.~\ref{fig:COT}(D), the phase shifts $\phi$ range from 0.32 to 0.68, with the COT decreasing as the phase shift increases from 0.32 to approximately 0.44, where it reaches its minimum of about 1.6, down from 2.7. Beyond this range, the COT stabilizes before slightly increasing at higher phase shifts around 0.60. With a phase shift smaller than 0.32, the robot tends to tilt backward, causing calculation failures in the QP algorithm. The mid-range phase shifts (0.44 to 0.52) exhibit lower variability, indicating more consistent and predictable energy efficiency. This suggests that these phase shifts are optimal for reducing energy costs at lower speeds. Moving beyond this range of phase shifts results in excessive pitching oscillations, leading to instability in the robot’s motion.

As for stride duration $T$, the COT decreases as the stride duration increases from 0.14 to around 0.24 seconds, dropping from 2.4 to 1.6. Shorter stride durations show higher variability and lower efficiency due to more touch-down events and collision losses, often leading to irregular foot sequences. Stride durations between 0.22 and 0.26 seconds provide more stable and energy-efficient outcomes. However, when the stride duration exceeds 0.34 seconds, it becomes difficult to stabilize the torso’s height and rotations, which can lead to slip motion during the stance phases and disrupt the planned footfall sequences.

\subsubsection{1.5 m/s} At mid-speeds, the feasible range of duty factor narrows, and fewer variations in COT are observed within this range. When the duty factor exceeds 0.60, there isn't enough time during the swing phase, causing instability in the robot. The COT decreases from approximately 2 at a duty factor of 0.18 to around 1.3 at 0.22, after which it stabilizes. The reduced variability at higher duty factors indicates that values above 0.22 result in more consistent and energy-efficient locomotion.

Compared to lower speeds, the variations in COT with phase shift become less pronounced at higher speeds. The COT decreases from around 1.7 to 1.5 as the phase shift increases from 0.36 to approximately 0.44. Beyond this point, the COT rises slightly, but mid-range phase shifts (0.44 to 0.52) continue to exhibit the lowest and most consistent values, indicating that this range optimizes energy efficiency at 1.5 m/s. Once the phase shift exceeds 0.56, the average COT remains constant. As shown in Fig.~\ref{fig:COT}(E), with larger phase shifts (above 0.60), the COT becomes less predictable as the robot's motion becomes increasingly unstable. With a phase shift smaller than 0.36, the robot tilts backward so severely that the optimization algorithm cannot find an appropriate solution.

In terms of stride duration, the COT decreases as the duration increases from 0.16 to 0.34 seconds, reaching a minimum of about 1.2 before stabilizing. Shorter stride durations once again exhibit greater variation, reinforcing that longer stride durations are more energy-efficient at this speed. However, when the stride time exceeds 0.34 seconds, significant torso pitch movements are observed, increasing the risk of foot slippage. Additionally, continuous simulation for more than 30 strides becomes difficult, and insufficient data can be gathered.

\subsubsection{2.5 m/s} At the top speed, the duty factor follows a similar trend (Fig.~\ref{fig:COT}(C)), with the COT decreasing from around 5 to 1.6 as the duty factor increases from 0.14 to 0.22. On the other hand, fewer solutions are found with duty factors greater than 0.50, as the absence of a flight phase significantly limits the stride length and maximal speeds.

Similarly, very few solutions are found with varying phase shifts. The COT is nearly constant, at approximately 1.5, with mid-range phase shifts (0.44 to 0.52) continuing to optimize energy efficiency. At high speeds, the torso is positioned very close to the ground, and small variations in $\phi$ can cause the torso to tilt, leading to premature ground contact by the legs. This instability results in immediate failure of the controller.

Compared to the COT at mid-speeds, the effect of stride duration follows a similar trend: the COT decreases as the stride duration increases from 0.16 to 0.34 seconds, dropping from around 2.2 to 1.2. Shorter stride durations continue to exhibit higher variability and inefficiency, while longer strides lead to more stable and energy-efficient locomotion. A similar failure pattern is observed when the stride duration exceeds 0.34 seconds.

\subsection{Absolute Power Analysis}
\begin{figure}[tbp]
\centering
\includegraphics[width=1\columnwidth]{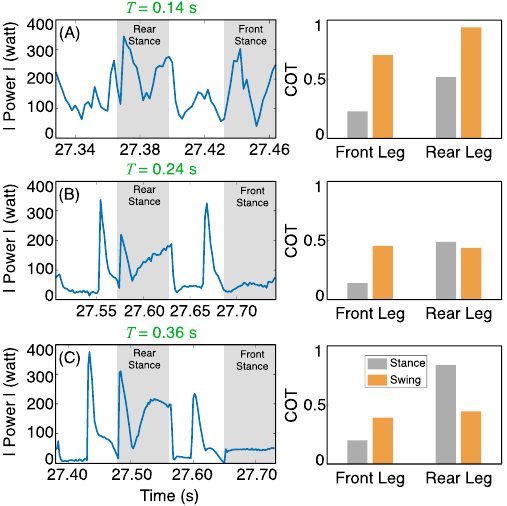}
\caption[Power]{This figure illustrates the absolute power consumption across all joints during a single stride at three different stride durations for the bounding gait at 0.5 m/s: (A) 0.14 seconds, (B) 0.24 seconds, and (C) 0.36 seconds.} 
\label{fig:Power}
\vspace{-2mm}
\end{figure}
To fully understand what caused the variations in the COT when varying the parameters, a case study on stride duration has been shown in this section.
Fig.~\ref{fig:Power} presents the absolute power consumption across all joints during a single stride of the bounding gait at 0.5 m/s. Three stride durations are investigated, as highlighted in Fig.~\ref{fig:COT}. In the case of the shorter stride duration (0.14 seconds, Fig.~\ref{fig:Power}(A)), power consumption fluctuates more frequently, with sharp peaks and valleys throughout the gait cycle. These rapid fluctuations reflect the increased mechanical work required to execute faster, shorter strides. Conversely, for the optimal stride duration (0.24 seconds, Fig.~\ref{fig:Power}(B)), power spikes are less frequent but more pronounced.
Bounding with a stride duration of 0.24 seconds is more energy-efficient. Comparing the case with shorter stride time, longer strides reduce the number of gait cycles per unit time, which in turn decreases the frequency of mechanical adjustments required by the actuators, particularly during the flight phases. This results in fewer rapid changes in power consumption, as seen in the longer stride duration case, where power consumption fluctuates less during the aerial phase. 
On the other hand, with excessive stride time (0.36 seconds as shown in Fig.~\ref{fig:Power}(C)), greater forces with longer durations are required during the rear stance phase. This inevitably increases energy consumption, as the robot moves away from its optimal oscillation frequency and remains in the air for a longer time.
\subsection{Hardware Validation}

\begin{figure}[tbp]
\centering
\includegraphics[width=1\columnwidth]{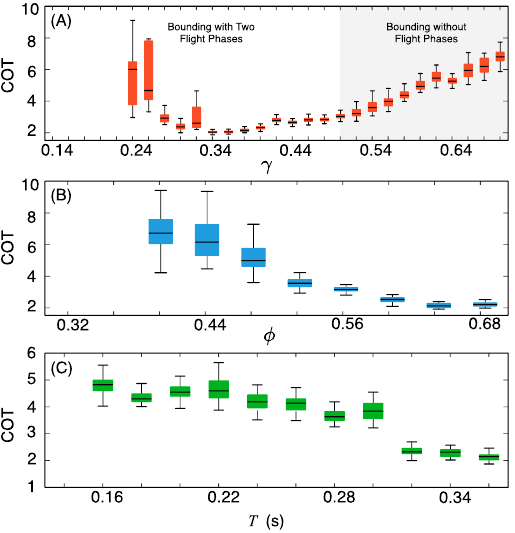}
\caption[ExpData]{These figures present experimental data of the COT for the A1 robot employing the bounding gait at a speed of 0.5 m/s, focusing on the three gait parameters. (A) illustrates the relationship between duty factor and COT, highlighting two distinct bounding gaits: bounding with two flight phases ($\gamma <$ 0.50) and bounding without flight phases ($\gamma >$ 0.50). Lower COT values are observed in the region with two flight phases. (B) illustrates the relationship between phase shift and COT, showing a decrease in COT with increasing phase shift values, indicating improved energy efficiency at higher phase shifts. (C) displays the influence of stride duration on COT, with results suggesting that longer stride durations generally correlate with lower COT, highlighting the energy efficiencies gained at longer strides.} 
\label{fig:ExpData}
\vspace{-2mm}
\end{figure}
To validate our simulation results, we conducted experimental tests on all three parameters at a speed of 0.5 m/s using the A1 robot on a treadmill. 
Fig.~\ref{fig:ExpData} illustrates the relationship between these parameters and the COT based on the experimental data. 
Despite model discrepancies such as the absence of motor dynamics, mechanical damping, and friction in the simulations, we observe similar trends when compared to the simulation data at the same speed (first column of Fig.~\ref{fig:COT}):
For the duty factor $\gamma$, as shown in Fig.~\ref{fig:ExpData}(A), the optimal value of approximately 0.30 is consistent with the simulation results. Deviating from this optimal value results in significantly higher energy consumption, increasing by up to four times.
Similarly, for phase shift $\phi$, smaller phase shifts result in higher COT values, while increasing the phase shift leads to more efficient gaits. The optimal phase shift observed from the hardware tests is approximately 0.64 (Fig.~\ref{fig:ExpData}(B)), compared to 0.50 in the simulation results. Additionally, the hardware tests show much larger variations (about four times) in COT when changing the phase shift.
Lastly, Fig.~\ref{fig:ExpData}(C) demonstrates that longer stride durations $T$ are associated with lower COT values, indicating improved energy efficiency. However, unlike the clear optimal stride time of 0.26 seconds observed in simulations, the experimental tests do not show the same values. With a stride time of 0.36 seconds, the hardware demonstrated the most efficient motion, though it was less stable due to excessive body rotations.

% This experimental data from the A1 robot validates our simulation findings, showing that adjustments in phase shifts and stride durations can significantly influence the energetic efficiency of the robot locomotion. In general, increasing phase shifts and extending stride durations both contribute to a decrease in COT at a consistent speed of 0.5 m/s. 

\section{Conclusions}
\label{sec:conclusions}
In this work, we investigated the impact of gait parameters (duty factor, phase shift, and stride duration) on the energy efficiency of quadrupedal locomotion using the Unitree A1 robot. Through comprehensive simulations and experimental validations, we demonstrated that controlling these gait parameters is \Review{crucial}{important}{1-2} for achieving efficient locomotion in legged robots. Our findings reveal that there is no unique selection of gait parameters; instead, the optimal combination varies with the robot's average speed. By systematically adjusting the gait parameters according to speed, we can significantly reduce the COT, enhancing the robot's performance and energy efficiency. This emphasizes the importance of adaptive gait control strategies that tailor locomotion patterns to the robot's operating conditions, paving the way for more practical and efficient quadrupedal robots in real-world applications.

{\appendix[Contact Jacobian]
\Review{}{The A1 robot feet are modeled as point contacts during stance phases since they are constructed from 0.02 $m$ rubber spheres. Let $\vec{p}_i = \left[ p_{i,x}(\vec{q}), p_{i,y}(\vec{q}), p_{i,z}(\vec{q}) \right]^\intercal \in \mathbb{R}^3$ represents the $i$-th foot position in the inertial frame as a function of the robot's generalized coordinates, $\vec{q}$, calculated using forward kinematics. Then, $\vec{J}_i = \frac{\partial \vec{p}_i(\vec{q})}{\partial \vec{q}} \in \mathbb{R}^{3 \times 18}$ is the corresponding contact Jacobian matrix. The vector of GRF acting on the foot is $\vec{\lambda}_i = \left[\lambda_{i,x}(t), \lambda_{i,y}(t), \lambda_{i,z}(t)\right]^\intercal \in \mathbb{R}^3$.}{8-1}

\bibliographystyle{IEEEtran}
\bibliography{References}

\end{document}